\documentclass[conference]{IEEEtran}
\IEEEoverridecommandlockouts
\usepackage{cite}
\usepackage{amsmath,amssymb,amsfonts}
\usepackage{algorithmic}
\usepackage{graphicx}
\usepackage{textcomp}
\usepackage{xcolor}
\usepackage{amssymb}
\usepackage{amsmath}
\usepackage{graphicx}
\usepackage{algorithm}
\usepackage{algorithmic}
\usepackage{subcaption}
\usepackage{newfloat}
\usepackage{times}  
\usepackage{helvet}  
\usepackage{courier}  
\usepackage{booktabs}
\usepackage{multirow} 
\usepackage[hyphens]{url}  
\usepackage{bm}
\usepackage{amsmath}
\usepackage{amssymb}
\usepackage{graphicx} 
\usepackage{listings}
\usepackage{color, xspace}
\newcommand{\tbtensor}[1]{\bm{\mathcal{\MakeUppercase{#1}}}}

\newcommand{\tbmatrix}[1]{\mathbf{\MakeUppercase{#1}}}
\def\BibTeX{{\rm B\kern-.05em{\sc i\kern-.025em b}\kern-.08em
    T\kern-.1667em\lower.7ex\hbox{E}\kern-.125emX}}

\begin{document}

\title{MetaLoRA: Tensor-Enhanced Adaptive  \\Low-Rank Fine-tuning }

\author{\IEEEauthorblockN{Maolin Wang}
Supervised by Prof. Xiangyu Zhao\\ Co-supervised by Dr. Ruocheng Guo and Prof. Junhui Wang\\
\IEEEauthorblockA{\textit{Department of Data Science}, 
\textit{City University of Hong Kong}\\
Hong Kong SAR, China \\
Morin.wang@my.cityu.edu.hk}}

\maketitle

\begin{abstract}
There has been a significant increase in the deployment of neural network models, presenting substantial challenges in model adaptation and fine-tuning. Efficient adaptation is crucial in maintaining model performance across diverse tasks and domains. While Low-Rank Adaptation (LoRA) has emerged as a promising parameter-efficient fine-tuning method, its fixed parameter nature limits its ability to handle dynamic task requirements effectively. Adapting models to new tasks can be challenging due to the need for extensive fine-tuning. Current LoRA variants primarily focus on general parameter reduction while overlooking the importance of dynamic parameter adjustment and meta-learning capabilities. Moreover, existing approaches mainly address static adaptations, neglecting the potential benefits of task-aware parameter generation in handling diverse task distributions. To address these limitations, this Ph.D. research proposes a LoRA generation approach to model task relationships and introduces MetaLoRA, a novel parameter-efficient adaptation framework incorporating meta-learning principles. This work develops a comprehensive architecture that integrates meta-parameter generation with adaptive low-rank decomposition, enabling efficient handling of both task-specific and task-agnostic features. MetaLoRA accurately captures task patterns by incorporating meta-learning mechanisms and dynamic parameter adjustment strategies. To our knowledge, this research represents the first attempt to provide a meta-learning enhanced LoRA variant, offering improved adaptation capability while maintaining computational efficiency in model fine-tuning.
\end{abstract}


\section{Introduction}

The scale of deep learning models has rapidly expanded, especially in natural language processing (NLP)~\cite{touvron2023llama,sengupta2022tensor}, Recommendation~\cite{zhao2018recommendations,zhao2018deep,wang2023multi,liu2020automated,liang2023mmmlp,li2022gromov} and computer vision (CV)~\cite{croitoru2023diffusion,hayashi2019exploring}. While pre-trained models like Llama~\cite{touvron2023llama} and diffusion models~\cite{croitoru2023diffusion} have achieved unprecedented performance, their practical deployment faces significant challenges. The enormous computational, temporal, and economic costs required for fine-tuning these models severely limit their accessibility and adaptability~\cite{han2024parameter}.

To address these challenges, parameter-efficient fine-tuning (PEFT) methods have emerged as promising solutions. Among them, Low-rank adaptation (LoRA)~\cite{hu2021lora} introduces low-rank decomposition matrices to modify pre-trained model weights, dramatically reducing the number of trainable parameters. This approach has demonstrated remarkable success across various large-scale models, with applications ranging from language understanding to generation tasks. Recent studies have shown that LoRA can achieve comparable performance to full fine-tuning while using only 0.1\%-1\% of the trainable parameters.

However, despite its success, LoRA faces critical limitations. The most significant challenge lies in the nature of fixed parameters, which fail to accommodate diverse data distributions and task requirements. This rigidity results in suboptimal performance and reduced generalization capability, particularly when dealing with tasks that differ significantly from the training distribution. To enhance LoRA's adaptability, several variants have been proposed~\cite{agiza2024mtlora,liu2023moelora,zadouri2023pushing}. For example, MTLoRA~\cite{agiza2024mtlora} introduces a dual-module architecture combining task-agnostic and task-specific components, while MOELoRA~\cite{liu2023moelora} leverages mixture of experts to handle different tasks through specialized embeddings.

Nevertheless, these variants still face two major challenges. First, there exists a significant performance gap compared to full fine-tuning, with accuracy differences of up to 5-10\% in complex tasks. Second, they exhibit limited dynamic adaptability due to their rigid architectural constraints, particularly when handling previously unseen task variations. These limitations underscore the urgent need for a more flexible and robust approach that can dynamically adjust parameters based on input characteristics while maintaining computational efficiency.
Given LoRA's inherent low-rank properties and parameter decomposability, tensor network methods~\cite{sengupta2022tensor,wang2023tensor} emerge as a particularly promising framework for creating adaptive LoRA structures. Tensor networks have demonstrated remarkable capabilities in modeling multi-task relationships~\cite{yang2016deep}. Their success in designing efficient CNNs~\cite{hayashi2019exploring} and facilitating cross-task collaboration~\cite{yang2016deep} suggests their potential for enhancing generalization capabilities while reducing computational overhead at the same time.

Recent advances in tensor computation and decomposition techniques have made it increasingly feasible to incorporate tensor networks into deep learning architectures. The ability of tensor networks to capture complex relationships while maintaining computational efficiency aligns perfectly with the goals of parameter-efficient adaptation. Moreover, tensor networks provide a theoretical foundation for understanding and controlling the trade-off between model expressiveness and computational complexity.

This Ph.D. research proposes \textbf{Meta}-enhanced \textbf{Lo}w-\textbf{R}ank \textbf{A}daptation (\textbf{MetaLoRA}), a novel framework that integrates meta-learning principles with LoRA. Our planned research methodology includes:

\begin{itemize}
    \item Developing a theoretical framework for incorporating tensor networks into LoRA architectures, with focus on maintaining parameter efficiency while enabling dynamic adaptation.
    
    \item Designing and implementing a meta-learning based parameter generation mechanism that can efficiently adjust LoRA parameters based on task characteristics.
    
    \item Investigating the trade-offs between model expressiveness and computational efficiency in the context of tensor-enhanced LoRA variants.
    
    \item Conducting some empirical studies to validate the effectiveness of the proposed approach across different tasks.
\end{itemize}

\section{Preliminaries and Background}
\begin{figure}[t]
\centering
\includegraphics[width=0.45\textwidth]{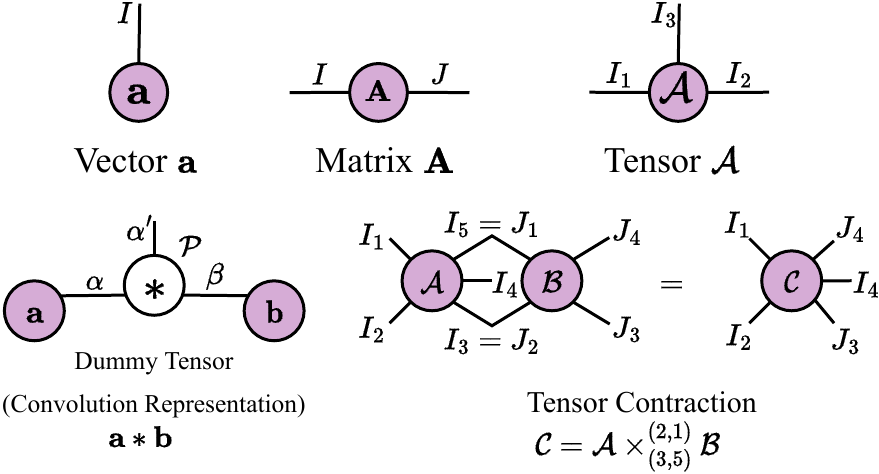}
\caption{Illustration of baisc tensor diagrams of a vector~(1st-order tensor), a matrix~(2rd-order tensor), a 3rd-order tensor, convolution representation and Tensor Contraction}
\label{fig:tensor_diagram}
\end{figure}

\subsection{Notations} 
In this paper, we use boldface Euler script letters to denote a $d$-order tensor, such as $\tbtensor{D} \in \mathbb{R}^{L_1 \times L_2 \times \cdots \times L_d}$. Each element of the tensor, with all subscripts specified, is represented as $\tbtensor{D}_{l_1, l_2, \ldots, l_d} \in \mathbb{R}$. By fixing a subset of the subscripts, we can extract a sub-tensor. For instance, fixing $L_1 = l_1$ and $L_2 = l_2$, we obtain a sub-tensor $\tbtensor{D}_{l_1, l_2} \in \mathbb{R}^{L_3 \times \cdots \times L_d}$. We represent vectors with bold lowercase letters, such as $\mathbf{v} \in \mathbb{R}^L$, and matrices with bold uppercase letters, such as $\mathbf{M} \in \mathbb{R}^{L_1 \times L_2}$. Vectors and matrices are treated as 1-order and 2-order tensors, respectively. As shown in Figure~\ref{fig:tensor_diagram}, the tensor diagrams illustrate the notation for 1st-order (vector), 2nd-order (matrix), and 3rd-order tensors.


\subsection{Tensor Contraction and Tensor Networks}

Tensor contraction, a fundamental operation in tensor networks, involves summing over common indices between two tensors, effectively reducing the overall dimensionality. Let ${\tbtensor{A}} \in \mathbb{R}^{I_1 \times I_2 \times \cdots \times I_N}$ and ${\tbtensor{B}} \in \mathbb{R}^{J_1 \times J_2 \times \cdots \times J_M}$ be tensors with shared indices such that $I_{n_k} = J_{m_k}$ for $k = 1, \dots, S$. The contraction of these tensors, denoted by ${\tbtensor{A}} \times_{(n_1, \dots, n_S)}^{(m_1, \dots, m_S)} {\tbtensor{B}}$, produces a tensor $\tbtensor{C}$ of order $(N + M - 2S)$. This operation is mathematically represented as:

\begin{align}
    \tbtensor{C} &= {\tbtensor{A}} \times_{(n_1, \dots, n_S)}^{(m_1, \dots, m_S)} {\tbtensor{B}}= \sum_{i_{n_1}, \dots, i_{n_S}} \tbtensor{A}_{i_{n_1}, \dots, *} \tbtensor{B}_{*, i_{m_1}, \dots}.
\end{align}
Tensor networks are formed by systematically contracting multiple tensors, where each tensor contraction reduces the network's overall dimensionality and complexity. This process involves performing successive contractions between pairs of tensors, ultimately capturing intricate relationships.


\subsection{Dummy Tensor and Convolution}
\begin{figure}[t]
    \centering
    \includegraphics[width=0.45\textwidth]{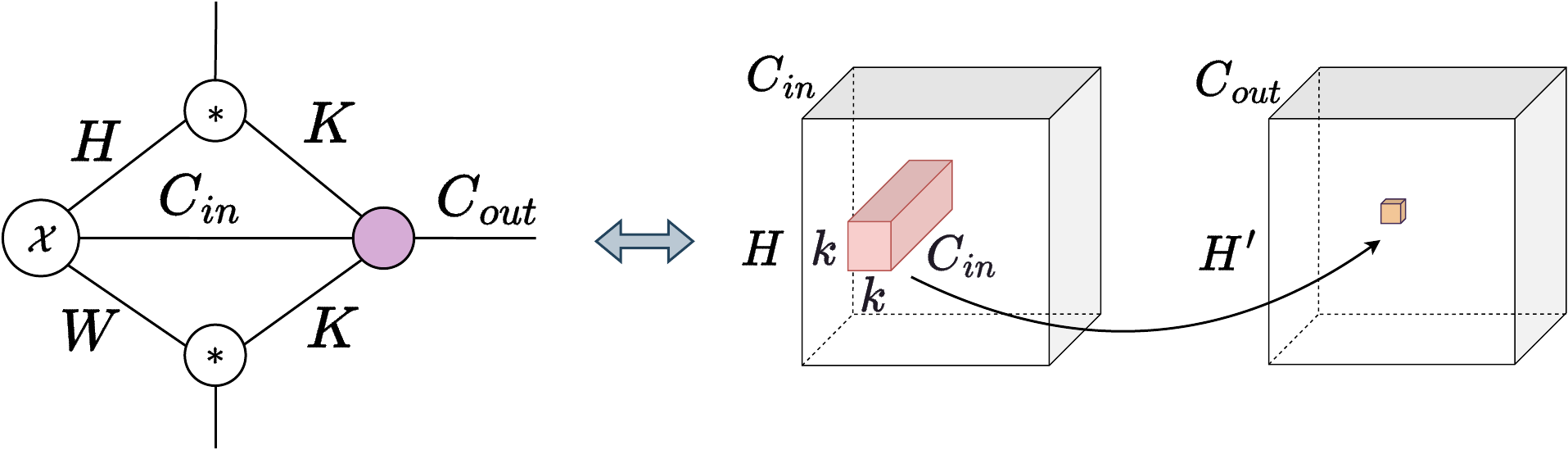}
    \caption{A tensor network representation of a simple convolutional layer, which two dummy tensors can represent to illustrate the most common image convolution. The convolution itself is a multilinear tensor operation.}
    \label{fig:tensor_network_convolution}
\end{figure}


As depicted in Figure~\ref{fig:tensor_diagram} and ~\ref{fig:tensor_network_convolution}, the dummy tensor~\cite{hayashi2019exploring} is shown as a node with star and arrow symbols. The convolution operation involving this tensor can be mathematically described as follows:

\begin{equation}
\label{eq:dummy}
\mathbf{y}_{j^{'}} =\mathbf{a}*\mathbf{b}
= \sum_{j=0}^{\alpha-1}\sum_{k=0}^{\beta-1} \tbtensor{P}_{j,j^{'},k}\mathbf{a}_{j} \mathbf{b}_{k},
\end{equation}

Here, $\mathbf{a} \in \mathbb{R}^{\alpha}$ denotes the input vector that is convolved with the weight vector $\mathbf{b} \in \mathbb{R}^{\beta}$, producing the output vector $\mathbf{y} \in \mathbb{R}^{\alpha '}$. 
The binary tensor $\tbtensor{P} \in \{0, 1\}^{\alpha \times {\alpha '} \times \beta}$ is defined such that $\tbtensor{P}_{j,j^{'},k}=1$ when $j = sj' + k - p$, and $0$ otherwise, where $s$ and $p$ refer to the stride and padding, respectively. This configuration allows $\tbtensor{P}$ to model the convolutional interactions between the tensors effectively.
Convolutional operations on tensors are crucial in various applications, especially in neural networks for image and signal processing. 

\subsection{CANDECOMP/PARAFAC and Tensor Ring Format}

The CANDECOMP/PARAFAC(CP) format~\cite{carroll1970analysis,kolda2009tensor} factorizes a higher-order tensor into a sum of rank-one tensors.
The Tensor Ring (TR) format~\cite{zhao2016tensor,sedighin2021adaptive} is introduced as a ring-shaped tensor network structure.
For an $N$th-order tensor $\tbtensor{X}\in\mathbb{R}^{I_1\times I_2\ldots I_N}$, the representation in CP format~\cite{kolda2009tensor} can be expressed as follows:
\begin{align}
\label{eq:pre:CP}
    \tbtensor{X}_{i_1,i_2,\ldots,i_N} &\approx \mathbf{\Lambda} \times^{1}_{1} \mathbf{A}^{(1)} \times^{2}_{1} \mathbf{A}^{(2)} \cdots \times^{N}_{1} \mathbf{A}^{(N)} \\
    &\approx \sum_{r=1}^R \lambda_r \prod^N_{n=1} \mathbf{A}^{(n)}_{i_n,r},
\end{align}
where $R$ represents the CP rank, $\tbmatrix{A}^{(n)}\in\mathbb{R}^{I_n\times R}$ are the factor matrices, and $\mathbf{\Lambda} = \text{diag}(\lambda_1, \ldots, \lambda_R)$ is a diagonal matrix containing the scaling factors that help in tuning the contributions of different rank-1 tensors.

\section{TECHNICAL IDEAS}
\begin{figure}[t]
    \centering
    \includegraphics[width=0.45\textwidth]{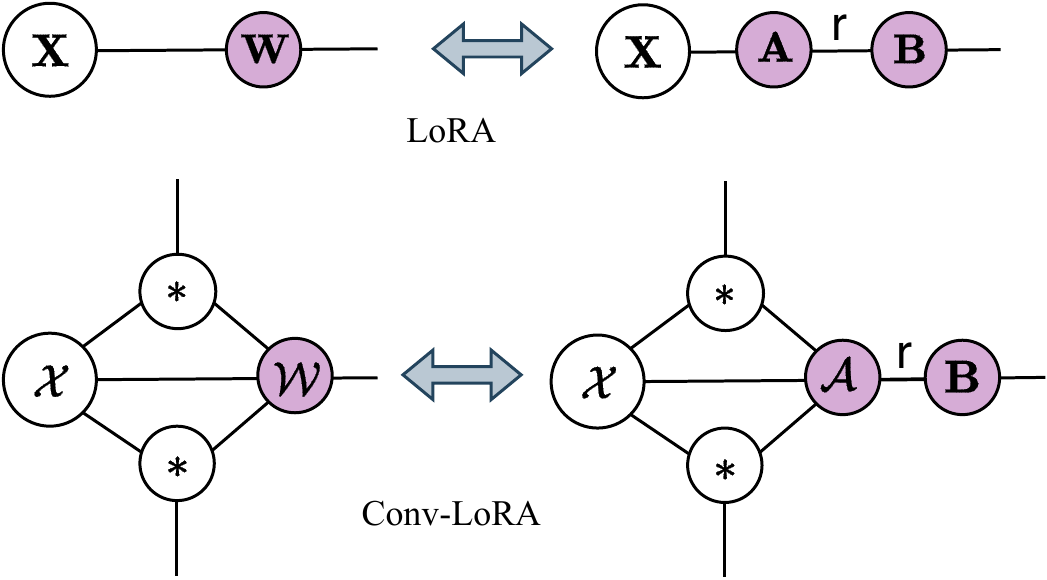}
    \caption{A tensor network representation of LoRA and our proposed Conv-LoRA. It can be observed that Conv-LoRA can be equivalently represented as a small convolution followed by a channel reduction (known as a 1$\times$1 convolution). The intuitive nature of tensor networks makes it easy to extend LoRA, which is applicable to matrices, to tensor operations.}
    \label{fig:conv_lora_tensor_network}
\end{figure}


This Ph.D. research introduces MetaLoRA, a novel method that extends the concept of Low-Rank Adaptation (LoRA) to tensor networks, specifically targeting convolutional neural networks. As illustrated in Figure~\ref{fig:conv_lora_tensor_network} and Figure~\ref{fig:tenta_lora_overview}, the architecture of MetaLoRA consists of three key modules: feature extraction, parameter space mapping net, and tensor-based parameter integration. The feature extraction module processes the input visual data, after which a mapping neural network generates parameters for model adaptation. Finally, these parameters are integrated using pre-defined tensor network structures, enabling efficient model adaptation.

\subsection{LoRA for Convolutional Networks}

Traditional LoRA implementations primarily focus on weight matrices~\cite{mao2024survey} and are not inherently suited for tensor operations characteristic of convolutional neural networks~\cite{he2016deep}. Our key contribution is extending LoRA to convolutional layers through tensor networks. As illustrated in Figure~\ref{fig:conv_lora_tensor_network}, the convolutional LoRA can be decomposed into a smaller convolution operation followed by a channel recovery process.

Specifically, for a convolutional tensor $\tbtensor{W} \in \mathbb{R}^{K\times K\times I \times O}$, where $K$ represents the spatial dimensions (kernel height and width), $I$ is the number of input channels, and $O$ is the number of output channels, this paper proposes the following \textbf{Conv-LoRA} formulation:
\begin{align}
\tbtensor{W}' = \tbtensor{W} + \Delta \tbtensor{W} = \tbtensor{W} + \tbtensor{A} \times^{4}_{1} \mathbf{B} = \tbtensor{W} + \sum_{r} \tbtensor{A}_{*r} \mathbf{B}_{r*}
\end{align}
Here, the tensor $\tbtensor{A} \in \mathbb{R}^{K\times K\times I \times R}$ represents the intermediate smaller convolutional filters. When the rank $R$ is small, the number of additional parameters introduced by this method is minimal, resulting in a highly efficient adaptation strategy. The matrix $\mathbf{B} \in \mathbb{R}^{R \times O}$ serves as the channel recovery matrix.

\subsection{Parameter Generation}

\subsubsection{Feature Extraction}
Feature extraction plays a pivotal role in MetaLoRA's effectiveness. The primary challenge lies in capturing the most relevant and high-level information from the input data. This paper utilizes pre-trained ResNet to obtain visual features from images, which are crucial for tasks requiring high spatial resolution and pattern recognition. The motivation behind this feature extraction process is to ensure that MetaLoRA can efficiently map features into the parameter space, facilitating dynamic adjustment based on the unique properties of the visual input data.

\subsubsection{Parameter Space Mapping Net}
Once extracted, the features must be effectively mapped from the sample space to the parameter space. This mapping is achieved using a multi-layer perceptron (MLP) that takes the feature vectors as input and outputs a set of parameters corresponding to the task. A specific challenge is ensuring that the MLP can generalize across different visual tasks while maintaining precision in parameter generation. The motivation behind this approach is to create a parameter generation mechanism that is versatile and efficient, capable of outperforming existing methods by offering greater flexibility and adaptability to new visual tasks.

\begin{figure}[t]
    \centering
    \includegraphics[width=0.45\textwidth]{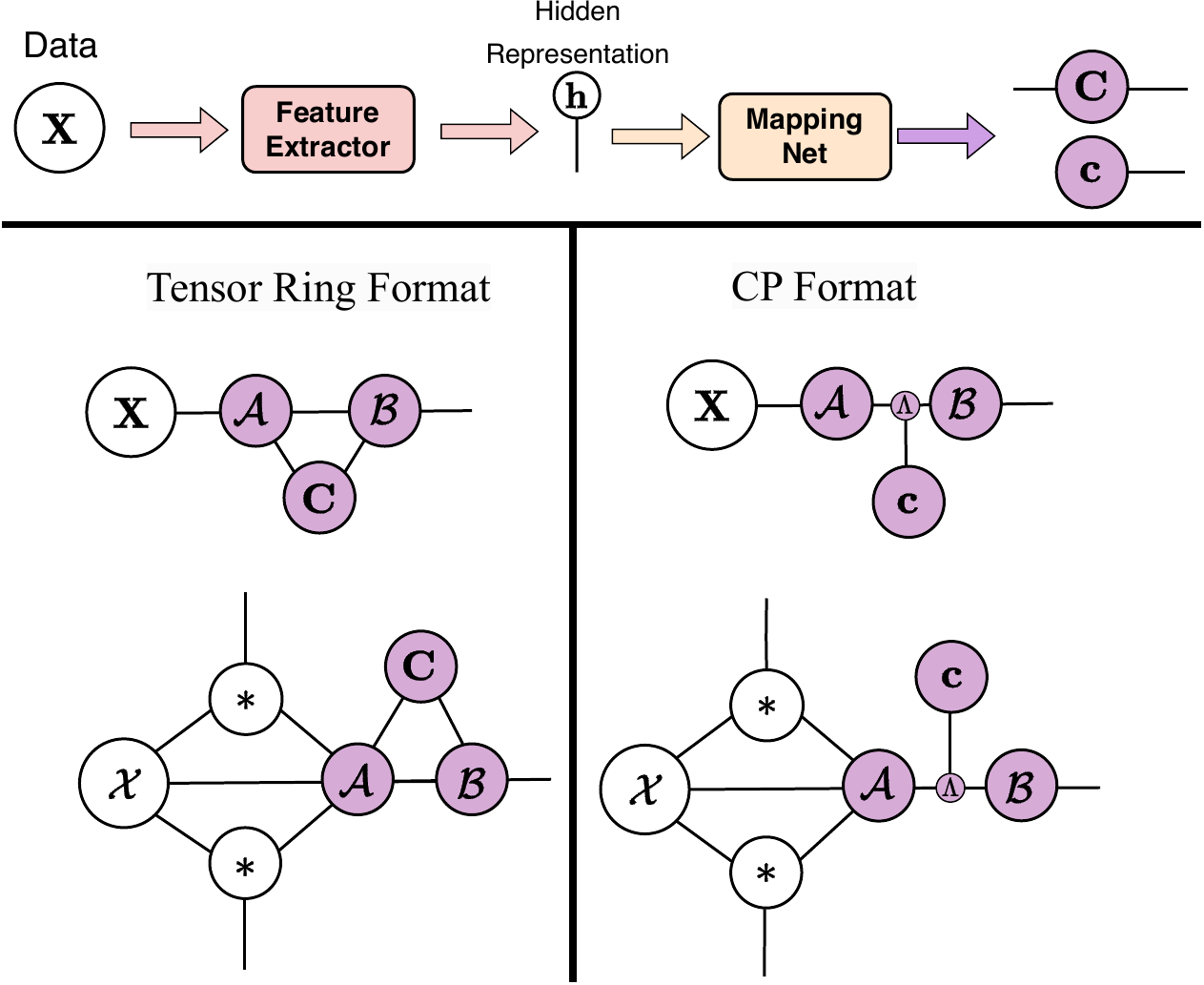}
    \caption{An overview of MetaLoRA's design. The figure illustrates the generation of the parameter $c$ by the mapping net, which is subsequently integrated into the model's weight matrices through CP and Tensor Ring format strategies. $\mathbf{\Lambda}$ is the identity diagonal tensor, meaning that all the elements on its diagonal are 1. The top section shows the generation of $\mathbf{c}$ or $\mathbf{C}$, while the bottom section demonstrates the application of the TR and CP formats to both general weight matrices and convolutional tensors.}
    \label{fig:tenta_lora_overview} 
    \vspace{-5pt}
\end{figure}

\subsection{Parameter Design of MetaLoRA}

The parameter design in MetaLoRA presents notable challenges, primarily in creating an integration strategy that ensures parameters are both contextually relevant and computationally efficient. After the mapping neural network generates the parameter seed $c$, these parameters are carefully integrated into the model's weight matrices. As illustrated in Figure~\ref{fig:tenta_lora_overview}, this process is formalized through two tensor network formats: CP format and tensor ring format.

For a general weight matrix $W$:
$$
\tbmatrix{W}' = \tbmatrix{W} + \Delta \tbmatrix{W},
$$ 

{MetaLoRA~(CP)} can be expressed as:
\begin{align}
\Delta \mathbf{W} &= \mathbf{\Lambda} \times^{1}_{1} \mathbf{A} \times^{2}_{1} \mathbf{B} \times^{3}_{1} \mathbf{c} = \sum_{r} \mathbf{A}_{*r} \mathbf{B}_{r*} \mathbf{c}_{r}
\end{align}
Here, $\mathbf{A}\in\mathbb{R}^{I\times R}$ and $\mathbf{B}\in \mathbb{R}^{R\times O}$ are factor matrices, and $\mathbf{c}_{r} \in \mathbb{R}^{R}$ represents the parameter generated via parameter space mapping net.

{MetaLoRA~(TR)} can be expressed as:
\begin{align}
\Delta \mathbf{W} &= \sum_{r_0,r_1,r_2} \tbtensor{A}_{r_0,*,r_1} \mathbf{B}_{r_1,*,r_2} \mathbf{C}_{r_2,r_0}
\end{align}
Here, $\tbtensor{A}\in\mathbb{R}^{R \times I\times R}$ and $\mathbf{B}\in \mathbb{R}^{R\times O \times R}$ are factor tensors, and $\mathbf{C}_{r} \in \mathbb{R}^{R \times R}$.

\subsection{Conv-LoRA's MetaLoRA Version}

For the MetaLoRA (CP) form for convolutional layers:
$$
\Delta \mathcal{W} = \mathbf{\Lambda} \times^{1}_{1} \mathcal{A} \times^{1}_{1} \mathbf{B} \times^{3}_{1} \mathbf{c} = \sum_{r} \mathcal{A}_{*r}  \mathbf{B}_{r*}  \mathbf{c}_r,
$$
where $\mathcal{A} \in \mathbb{R}^{K \times K \times I \times R}$ and $\mathbf{B} \in \mathbb{R}^{R \times O}$ are factor tensors, and $\mathbf{c}_r \in \mathbb{R}^{R}$ is a parameter generated via parameter space mapping network.

For the MetaLoRA (TR) form for convolutional layers:
$$
\Delta \mathbf{W} = \sum_{r_0, r_1, r_2} \mathcal{A}_{r_0, *, r_1} \mathcal{B}_{r_1, *, r_2} \mathbf{C}_{r_2, r_0},
$$
where $\mathcal{G}^{(1)} \in \mathbb{R}^{R \times I \times R}$ and $\mathbf{B} \in \mathbb{R}^{R \times O \times R}$ are factor tensors, $\mathcal{C} \in \mathbb{R}^{R \times R}$ represents a parameter matrix generated through the mapping network.

\subsection{Discussions of MetaLoRA}

Beyond the convolutional domain, MetaLoRA's principles could potentially be extended to other domains like large language models~\cite{xu2024multi}. The meta-learning nature of MetaLoRA makes it particularly suitable for personalized applications, such as recommendation systems where models need to adapt to individual user preferences. Furthermore, while this research focused on convolutional networks in this work, the framework's theoretical foundations in tensor networks suggest broader applications in transformer architectures and multimodal learning.
\section{Preliminary Results}

\begin{table}[t]
\caption{Model performance comparison for ResNet and MLP-Mixer architectures under different settings. {In this context, the LoRA of ResNet is implemented using our proposed Meta-LoRA. ``\textbf{{\Large *}}'' indicates the statistically significant improvements (i.e., two-sided t-test with $p<0.05$) over the best baseline.}}
\label{tab:model_performance_2}
\resizebox{0.5\textwidth}{!}{%
\begin{tabular}{l||c|c||c|c}
\hline
\textbf{Method} & \multicolumn{2}{c||}{\textbf{ResNet}} & \multicolumn{2}{c}{\textbf{MLP-Mixer}} \\
\hline
K in KNN  & \textbf{K=5} & \textbf{K=10} & \textbf{K=5} & \textbf{K=10} \\
\hline
Original        & 67.04\% & 61.36\% & 58.27\% & 60.83\% \\
LoRA            & 67.85\% & 62.02\% & 59.16\% & 61.22\% \\
Multi-LoRA      & 72.11\% & 68.57\% & 63.74\% & 65.49\% \\
\hline
Meta-LoRA CP   & 71.07\% & 71.29\% & 70.32\% & 72.52\% \\
Meta-LoRA TR   & 73.24\%{*} & 71.26\% & 71.75\%{*} & 73.87\%{*} \\
\hline
\end{tabular}%
}
\end{table}

Our preliminary experiments show the promising potential of Meta-LoRA across different visual architectures. As shown in Table~\ref{tab:model_performance_2}, Meta-LoRA TR demonstrates consistent improvements over baseline methods. The results reveal several key insights: First, the tensor-based parameter generation approach shows strong generalization capabilities, particularly with the TR variant achieving up to 73.87\% accuracy on MLP-Mixer. Second, the integration of Meta-LoRA with ResNet architecture proves effective, suggesting that our approach successfully extends parameter-efficient fine-tuning to convolutional networks while preserving their structural properties. These initial findings suggest that exploring more sophisticated tensor network architectures could lead to further improvements in adaptation efficiency and effectiveness.

\section{Related Works}

\noindent\textbf{LoRA and LoRA Variants:}
Recent parameter-efficient fine-tuning (PEFT) techniques, such as Adapter Tuning~\cite{zhang2023llama}, Prefix-Tuning~\cite{li2021prefix}, and Low-Rank Adaptation (LoRA)\cite{mao2024survey}, have been developed to reduce the computational cost of adapting Large Language Models (LLMs) to new tasks. LoRA, in particular, has gained significant popularity by injecting trainable low-rank matrices into model layers, achieving near-full fine-tuning performance with substantially fewer computational resources and memory requirements. Advanced variants like MultiLoRA\cite{wang2023multilora} and Mixture-of-Experts LoRA (MOELoRA)~\cite{liu2023moelora} aim to improve model adaptability and task-specific performance but often struggle with cross-task generalization and knowledge transfer. This fundamental limitation motivates our development of Meta-LoRA, which leverages tensor networks for dynamic parameter adjustment and optimization. 

\noindent\textbf{Tensor Networks and Neural Networks:}
Tensor networks (TNs) have fundamentally advanced neural networks by enabling efficient model compression and parameter sharing across network layers~\cite{cichocki2016tensor}. Advanced techniques like Canonical Polyadic (CP) decomposition and Tucker decomposition effectively reduce model size and computational overhead, particularly crucial in resource-constrained mobile applications and edge computing scenarios~\cite{wang2023tensor}. Tensor Ring decomposition has been particularly effective in replacing traditional fully connected layers, achieving significant parameter reductions without sacrificing model accuracy or performance~\cite{xie2024neural}. Tensor networks have also shown remarkable promise in embedding compression and parameter reduction within Large Language Models, substantially improving memory efficiency and cross-task transferability~\cite{jie2023fact}. 

\section{Discussion}

This doctoral research proposes MetaLoRA as a potential pathway to address the fundamental limitations in contemporary parameter-efficient fine-tuning methods. While the framework suggests that tensor network integration could enable dynamic adaptation without compromising computational efficiency, significant challenges remain in bridging theoretical constructs with practical implementations. The preliminary mathematical analysis indicates that tensor decomposition properties may preserve LoRA's low-rank characteristics, but this hypothesis requires rigorous validation through more empirical studies. Early-stage experiments show modest accuracy improvements, yet these results remain inconclusive regarding the framework's generalizability across diverse task domains.
A critical open question centers on maintaining the delicate balance between enhanced adaptability and preserved parameter efficiency. 
\section*{Acknowledge}

This research was partially supported by Research Impact Fund (No.R1015-23), Collaborative Research Fund (No.C1043-24GF), APRC - CityU New Research Initiatives (No.9610565, Start-up Grant for New Faculty of CityU), Hong Kong ITC Innovation and Technology Fund Midstream Research Programme for Universities Project (No.ITS/034/22MS), Huawei (Huawei Innovation Research Program), Tencent (CCF-Tencent Open Fund, Tencent Rhino-Bird Focused Research Program), Ant Group (CCF-Ant Research Fund), Alibaba (CCF-Alimama Tech Kangaroo Fund No. 2024002), and Kuaishou.
\bibliographystyle{ieeetr}
\bibliography{sample}

\end{document}